\documentclass[conference]{IEEEtran}
\IEEEoverridecommandlockouts
\usepackage{cite}
\usepackage{amsmath,amssymb,amsfonts}
\usepackage{algorithmic}
\usepackage{graphicx}
\usepackage{subfig}
\usepackage{textcomp}
\usepackage{array}
\usepackage{xcolor}
\usepackage{layouts}
\def\BibTeX{{\rm B\kern-.05em{\sc i\kern-.025em b}\kern-.08em
    T\kern-.1667em\lower.7ex\hbox{E}\kern-.125emX}}
\begin{document}

\title{The inD Dataset: A Drone Dataset of Naturalistic Road User Trajectories at German Intersections
}

\author{Julian Bock$^{1}$, Robert Krajewski$^{1}$, Tobias Moers$^{2}$, Steffen Runde$^{1}$, Lennart Vater$^{1}$ and Lutz Eckstein$^{1}$
\thanks{$^{1}$The authors are with the Automated Driving Department, Institute for 
Automotive Engineering RWTH Aachen University (Aachen, Germany). \newline(E-mail: \{bock, krajewski, steffen.runde, vater, eckstein\}@ika.rwth-aachen.de).}
\thanks{$^{2}$The author is with the Automated Driving Department, fka GmbH (Aachen, Germany). (E-mail: tobias.moers@fka.de).}}

\let\oldtwocolumn\twocolumn
\renewcommand\twocolumn[1][]{%
    \oldtwocolumn[{#1}{
    \begin{center}
           \includegraphics[width=\textwidth]{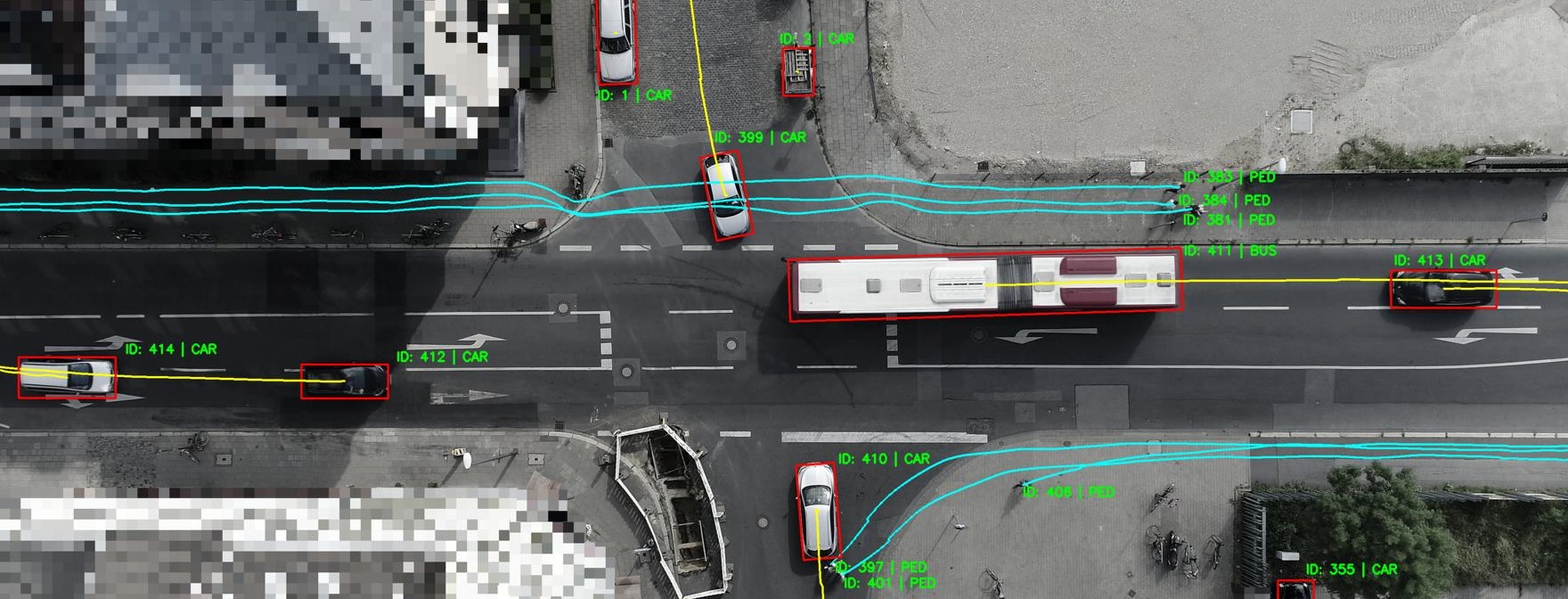}
           \captionof{figure}{Exemplary result of road user trajectories in the inD dataset. The position and speed of each road user is measured accurately over time and shown by bounding boxes and tracks. For privacy reasons, the buildings were made unrecognizable.}
           \label{fig:headliner}
        \end{center}
    }]
}

\maketitle
\begin{abstract}
Automated vehicles rely heavily on data-driven methods, especially for complex urban environments. Large datasets of real world measurement data in the form of road user trajectories are crucial for several tasks like road user prediction models or scenario-based safety validation.
So far, though, this demand is unmet as no public dataset of urban road user trajectories is available in an appropriate size, quality and variety. By contrast, the highway drone dataset (highD) has recently shown that drones are an efficient method for acquiring naturalistic road user trajectories. Compared to driving studies or ground-level infrastructure sensors, one major advantage of using a drone is the possibility to record naturalistic behavior, as road users do not notice measurements taking place.
Due to the ideal viewing angle, an entire intersection scenario can be measured with significantly less occlusion than with sensors at ground level. 
Both the class and the trajectory of each road user can be extracted from the video recordings with high precision using state-of-the-art deep neural networks. 
Therefore, we propose the creation of a comprehensive, large-scale urban intersection dataset with naturalistic road user behavior using camera-equipped drones as successor of the highD dataset.
The resulting dataset contains more than 11500 road users including vehicles, bicyclists and pedestrians at intersections in Germany and is called inD. 
The dataset consists of 10 hours of measurement data from four intersections and is available online for non-commercial research at: http://www.inD-dataset.com
\end{abstract}


\section{Introduction}
Automated driving is expected to reduce the number and severity of accidents significantly \cite{RoesenerImpact}. 
However, intersections are challenging for automated driving due to the large complexity and variety of scenarios \cite{intersectionChallenging}.
Scientists and companies are researching how to technically handle those scenarios by an automated driving function and how to proof safety of these systems.
An ever-increasing proportion of the approaches to tackle both challenges are data-driven and therefore large amounts of measurement data are required.
For example, recent road user behaviour models, which are used for prediction or simulation, use probabilistic approaches based on large scale datasets \cite{JulianBock.2017b, ridel2018literature}.
Furthermore, current approaches for safety validation of highly automated driving such as scenario-based testing heavily rely on large-scale measurement data on trajectory level \cite{highD2018, Wachenfeld.2016, ENABLES3.}.

However, the widely used ground-level or on-board measurement methods have several disadvantages. 
These include that road users can be (partly) occluded by other road users and do not behave naturally as they notice being part of a measurement due to conspicuous sensors \cite{highD2018}.
\begin{figure}[b!]
\centerline{\includegraphics[width=0.5\textwidth]{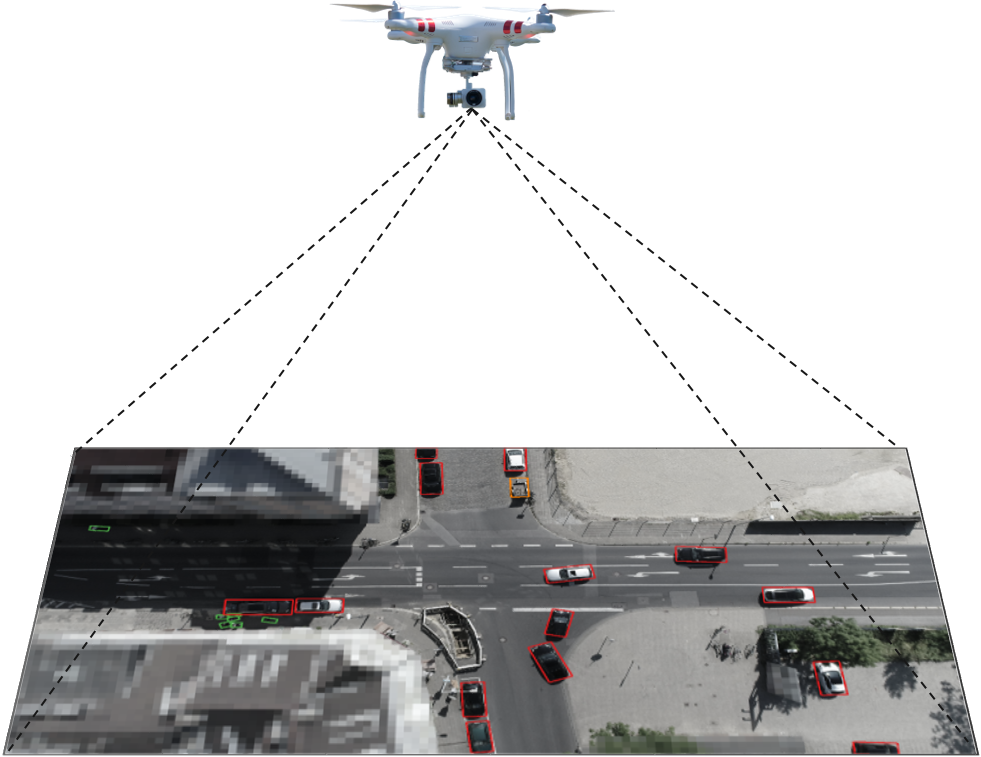}}
\caption{We propose to use a camera-equipped drone to record traffic at urban intersections. The trajectory of each road participant is accurately extracted using deep-learning algorithms.}
\label{fig:droneMethod}
\end{figure}
We propose to use camera-equipped drones to record road user movements at urban intersections (see Fig.~\ref{fig:droneMethod}).
Drones with high-resolution cameras allow to record traffic from a so-called bird's-eye perspective with high longitudinal and lateral accuracy.
The position and orientation of the objects is sufficient for most applications such as the prediction of road users.
Although information about the height of the objects is lost, it can be estimated from the class of the object, if necessary. 
Another advantage of the bird's-eye perspective is that occlusions by road users do not occur.
Finally, the recorded traffic behaviour is natural, as the drone, hovering at a height of up to 100~meters, is typically not perceived.\\   

In this paper, we apply methods similar to the approach in \cite{highD2018} to create a naturalistic road user trajectory dataset of German intersections called inD (\textbf{in}tersection \textbf{D}rone dataset). Exemplary road user tracks are shown in Fig.~\ref{fig:headliner}.
Also, we compare the inD dataset with other datasets that are commonly used in research, such as the Stanford Drone Dataset \cite{Robicquet.2016}.
With the publication of the dataset, we want to foster research on safety validation for highly automated driving, traffic simulation models, traffic analysis, driver models, road user prediction models and further topics, which rely on naturalistic traffic trajectory data from intersections. 
We use the common term drone throughout the paper for an unmanned aerial vehicle, which in our case is a multicopter.

\section{Related Work}
Within the last few years several datasets have been published which contain trajectories of road users, especially pedestrians. 
Nevertheless, none of the released datasets are suitable to solve open problems in the domain of automated driving, which we explain in the following. 
Subsequently, we give an overview of the datasets best comparable with inD, based on the recording technique used or the scenarios recorded.\\

Some of the first public datasets of road user trajectories, which are established in research, are the BIWI Walking Pedestrians datasets \cite{SPellegrini.2009} and the Crowds UCY/Zara dataset \cite{Lerner.2007}. The BIWI Walking Pedestrians consists of two datasets, which are the ETH and HOTEL dataset. The ETH dataset was recorded from the top of the ETH Zurich main building using a camera, while the HOTEL dataset was captured from top of a hotel. Although these datasets were used for a lot of research in the past, the datasets are of small size and contain no road users other than pedestrians.

The Stanford Drone Dataset \cite{Robicquet.2016} was the first trajectory dataset utilizing drones for recording road users' movements.
The dataset contains 10~300 trajectories of pedestrians, bicyclists, cars, skateboarders, carts and busses in the measured university campus area. 
Since pedestrians, bicyclists and skaters are most frequently represented, they comprise 94.5~\% of the trajectories.
Vehicles, however, represent only a very small proportion of road users in the dataset.

In 2019, two further similar datasets using drones were published, namely the CITR and DUT \cite{Yang2019TopviewTA}, which focused on the interaction between pedestrians and vehicles.
Like the Stanford Drone dataset, the DUT dataset was collected on the campus of a university.
One location is a shared space for all kinds of road users and the other location includes a pedestrian crosswalk at an unsignalized intersection.
While the DUT dataset contains uninstructed naturalistic trajectories of pedestrians, the CITR dataset is a controlled experiment on a parking lot. In this study, subjects were instructed to cross the way of a golf cart driving back and forth. 

\begin{figure*}[htb]
\subfloat[][Bendplatz, Aachen]{{\includegraphics[width=3.48761in]{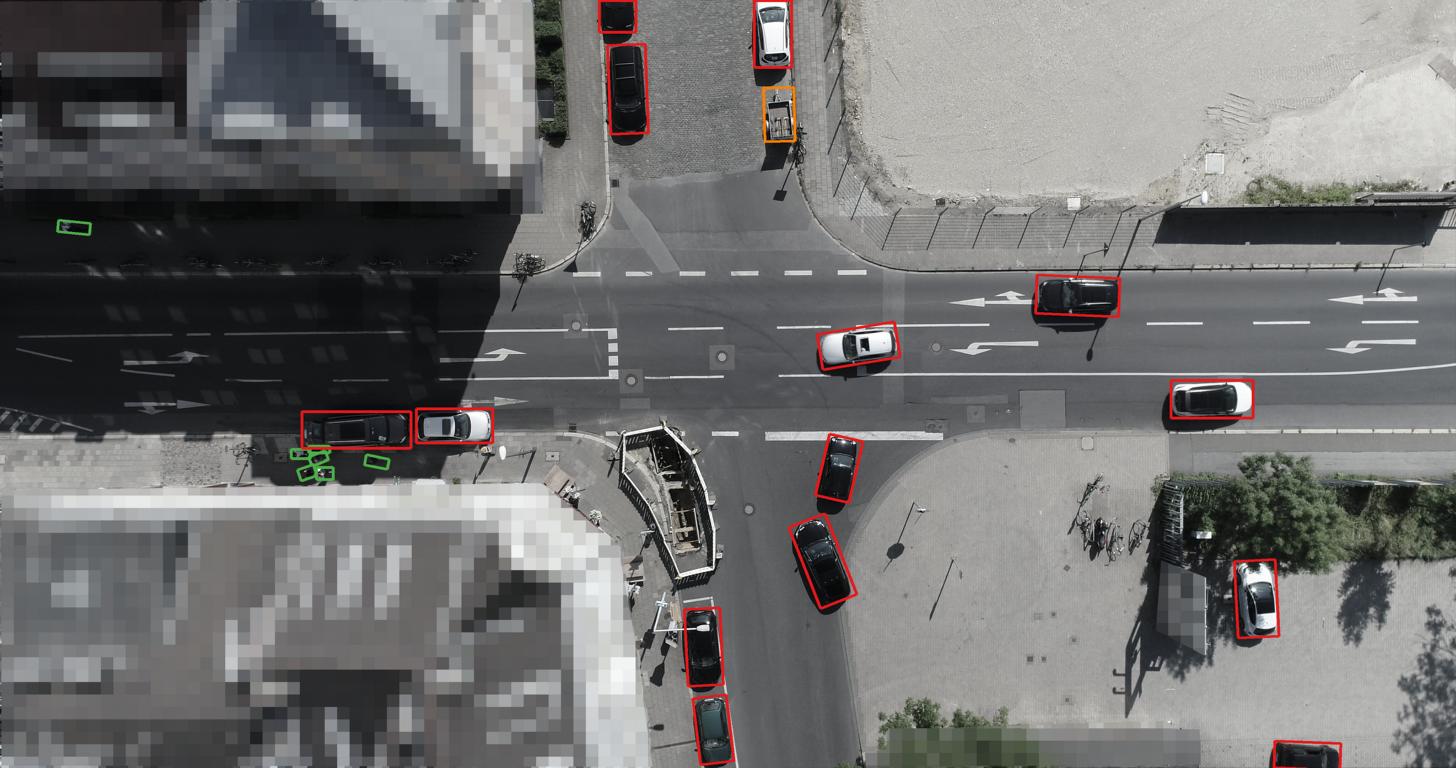} }}
	\quad
\subfloat[][Frankenburg, Aachen]{{\includegraphics[width=3.48761in]{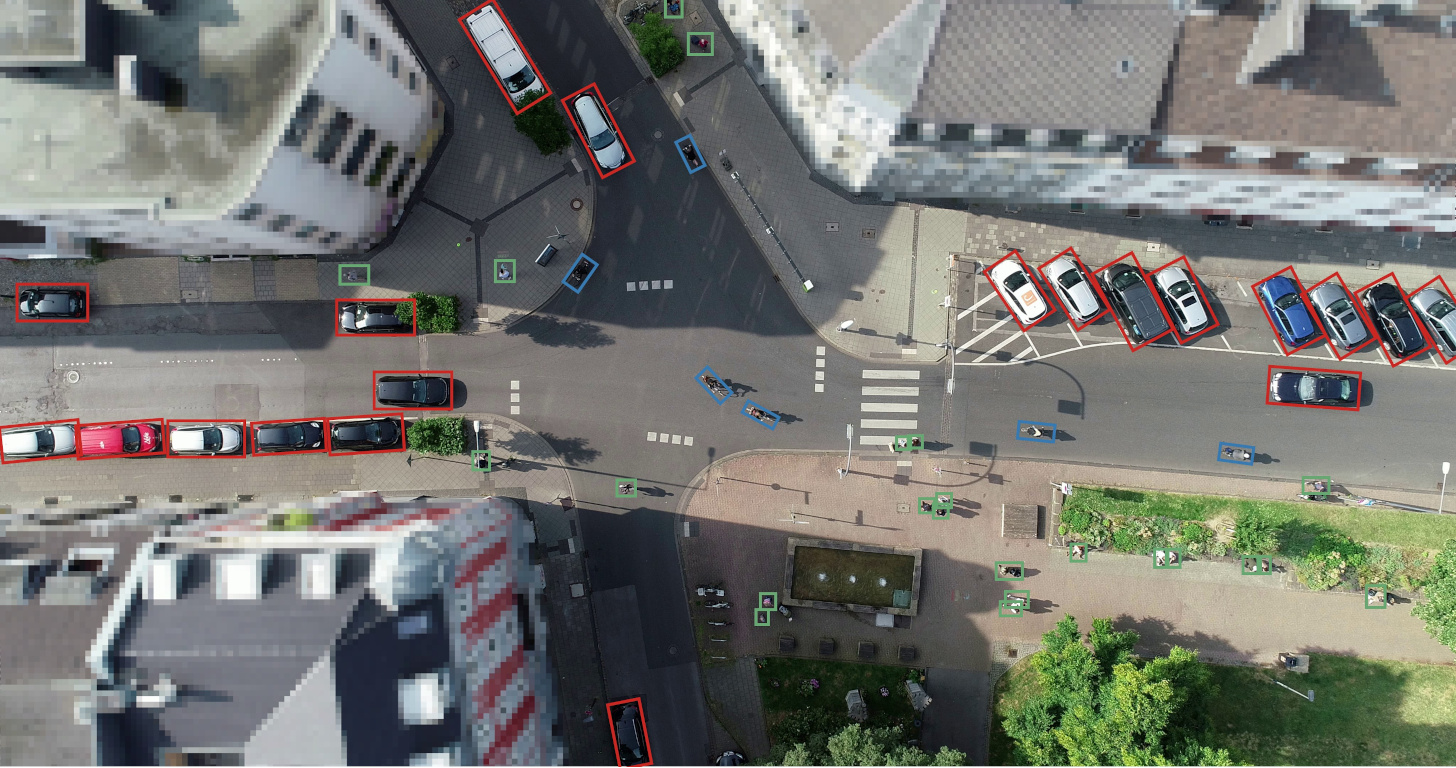} }}
	\quad
\subfloat[][Heckstrasse, Aachen]{{\includegraphics[width=3.48761in]{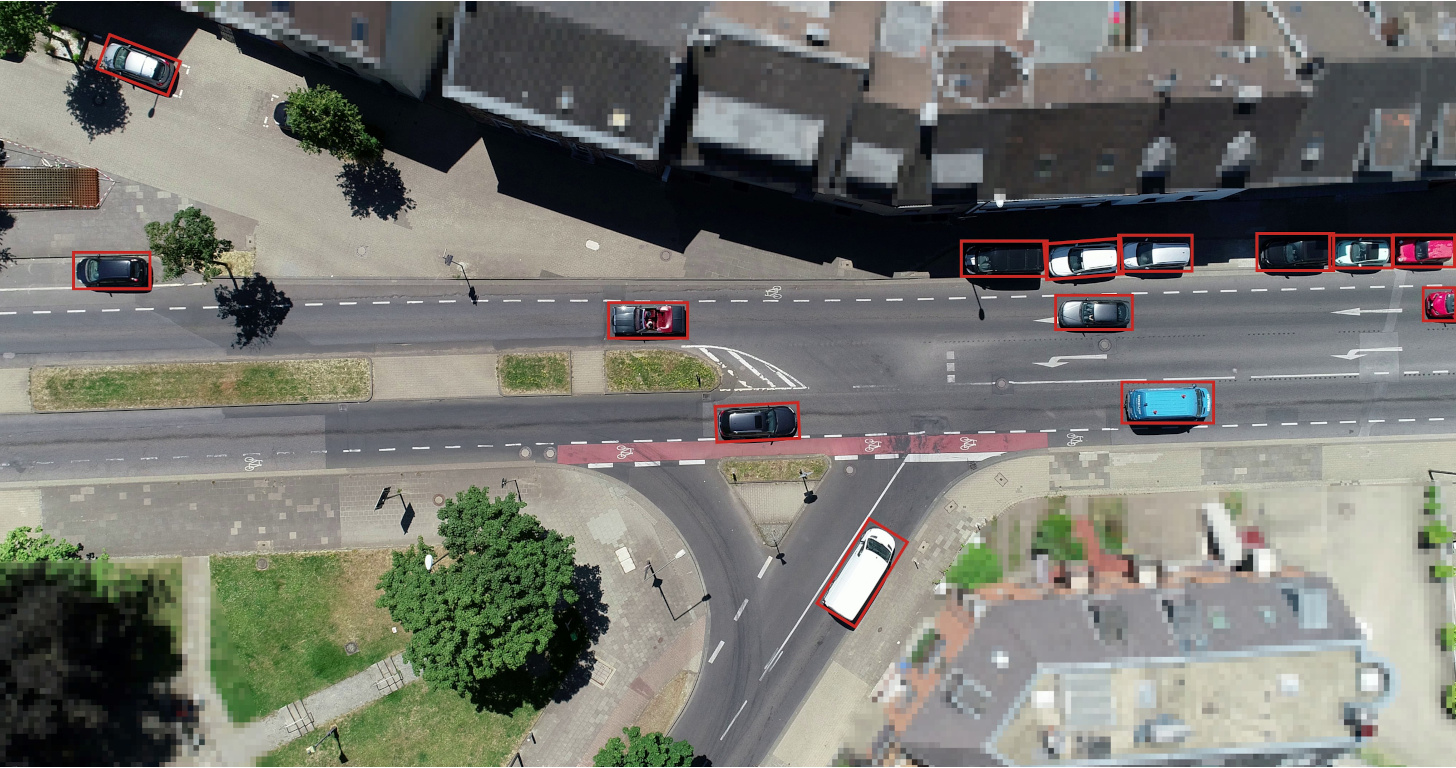} }}
	\quad
\subfloat[][Neuk\"ollner Strasse, Aachen]{{\includegraphics[width=3.48761in]{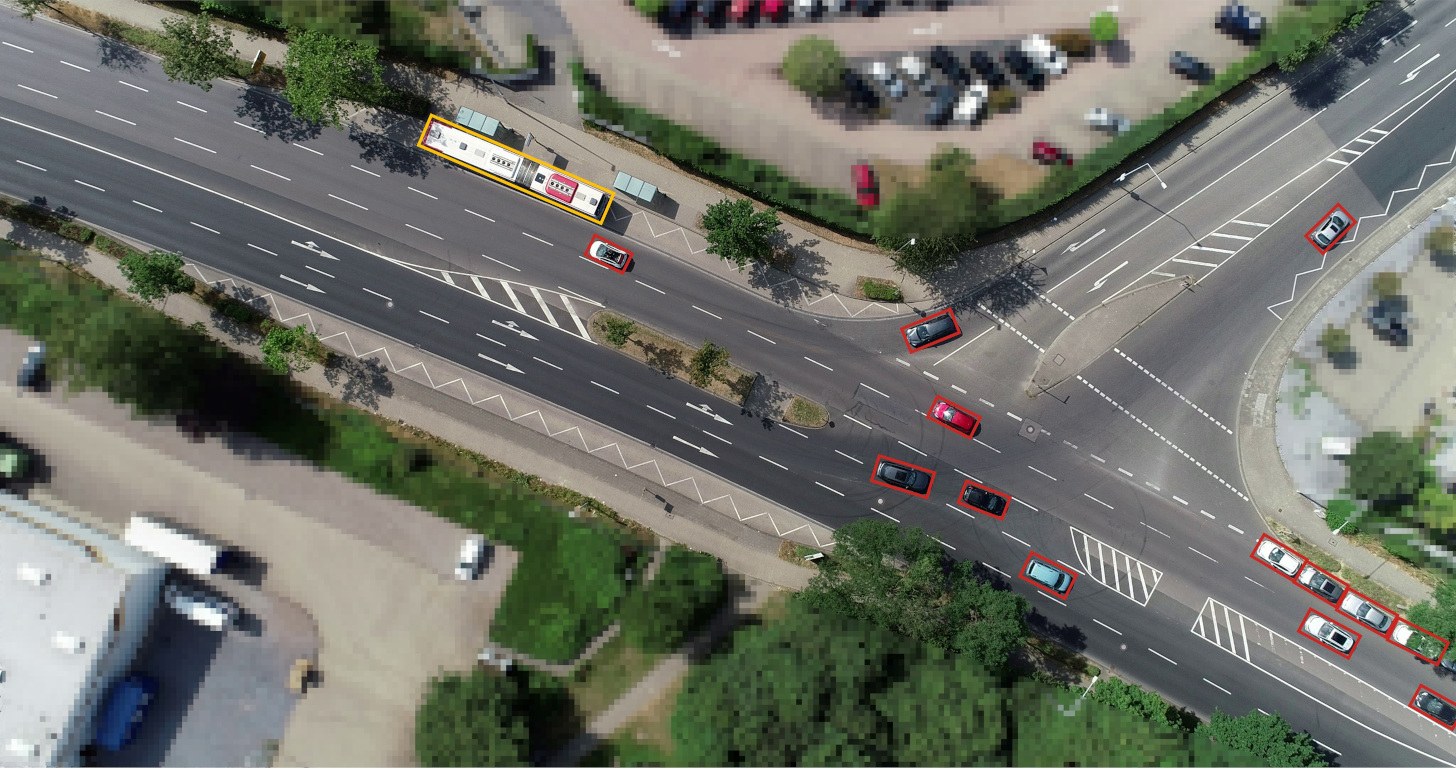} }}
\caption{Example images of four recording sites included in the inD dataset. Coloured bounding boxes show detected of road users (red: cars, orange: buses/trailers, green: pedestrians, blue: cyclists).}
\label{4fig}
\end{figure*}

With the highD dataset \cite{highD2018}, the first large-scale naturalistic vehicle trajectory dataset measured on public roads utilizing drones was published. The highD dataset contains 110~500 vehicles with a total travelled distance of 44~000 kilometers on German highways, measured with a positioning error of typically below 10~centimeters. The measurements took place at six different locations. However, the dataset was measured on highways and thus did not contain any Vulnerable Road Users (VRU).

The Ko-PER \cite{Strigel.2014} dataset was recorded at an intersection in Germany using stationary laser scanners and cameras. The dataset contains pedestrians, bicyclists, cars and trucks with in total 340~trajectories extracted from less than one hour of recordings.
From the same intersection, a new dataset called VRU Trajectory Dataset was later published \cite{goldhammer2018intentions}. The original VRU Trajectory Dataset consists of 1068~pedestrian and 464~cyclist trajectories, but was extended by the Extended Cyclist Trajectory Dataset with 1746 further cyclist trajectories. In the following, we consider both datasets combined under the title VRU Trajectory dataset.

Many of the mentioned datasets (Crowds, Ko-PER, CITR, BIWI Hotel) contain less than 1000~trajectories of all types of road users in total, which is not sufficient for most research on automated driving today. 
The Stanford Drone dataset contains about 10~000 trajectories making it more suitable for data-driven applications. 
However, only a small proportion of the road users in the dataset are vehicles and the dataset was created on a university campus, on which the road user behavior differs from public roads. The VRU Trajectory Dataset also does not contain any vehicle trajectories.
While this makes these datasets appropriate for investigating the interaction between pedestrians and bicyclists, they are not suitable for applications in the field of automated driving on public roads.

\section{Requirements for a trajectory dataset}
Based on the findings in \cite{highD2018}, we formulate the following requirements for a trajectory dataset of road users at urban intersections, which is appropriate for the application in data-driven methods in the field of automated driving:
\begin{itemize}
\item \textbf{Preserve the naturalistic behavior of road users}\\
Road users must not be influenced by the measurement method, e.g. by visible sensors that look like traffic surveillance cameras.
\item \textbf{Have a sufficient size} \\
The dataset must contain trajectories of several thousand road users. A sufficient size and variety is crucial for the applicability of the dataset for data-driven algorithms.
\item \textbf{Vary recording sites and times} \\
The dataset must include measurements from multiple recording sites at different times of the day to cover a variety of road layouts, traffic rules, and traffic densities. 
For applicability of the results in the field of automated driving on public roads, the measurements should mainly be carried out on public roads instead of private grounds.
\item \textbf{Detect and track all types of road users} \\
The dataset must not be limited to a certain road user category such as pedestrians or cars. All road users must be tracked as they influence each other.
\item \textbf{Track road users with high accuracy} \\
The resulting trajectories must have a positioning error of less than 0.1~meters independent of the road user type.
\item \textbf{Include the infrastructure}\\
Since the behaviour of road users is strongly dependent on the road layout and local traffic rules, both must be recorded precisely and provided in the dataset.
\end{itemize}

\section{Method}
In order to create a trajectory dataset using a drone, in addition to the recordings themselves further processing steps are necessary. 
Besides pre-processing the video recordings, this mainly includes the accurate automated detection, tracking and classification of all road users using computer vision algorithms. 
Manual annotation of the recorded road user trajectories would not be feasible, since the annotation for the extraction of the trajectories would have to be carried out in each video frame. 
The following sections give an overview of all relevant steps applied for creating the dataset.

\subsection{Selection of Recording Sites and Flight Approval}
After a total of four locations had been selected on the basis of traffic volume and traffic composition (see Fig.~\ref{4fig}), a safety concept was developed for each location individually. 
This included the selection of the landing site as well as the exact position of the drone during the recording. 
Thereby, not only the area covered during the recording had to be optimized, but local laws and maps specifying safety distances as well as no-fly zones also had to be taken into account.

\subsection{Recordings and Pre-processing}
Prior to each recording, the weather conditions were checked. 
To maximize recording quality, all recordings were taken in sufficient lighting conditions and with low winds.
This enhanced the sharpness and steadiness of the images, which facilitated further processing.
For each recording, the drone hovered at a pre-defined position at an altitude of up to 100~meters in order to completely cover the relevant intersection area.
The videos were recorded with a DJI Phantom 4 Pro in 4K (4096x2160~pixel) resolution at 25~frames per second in maximum video quality.
For each flight, we typically achieved a video recording time of approximately 20-22~minutes.
Although the drone uses flight stabilization control and gimbal-based camera stabilization during recording, translational and rotational motion could not be completely avoided. 
Therefore, in addition to correcting the lens distortion, each recording was also stabilized onto the first frame.

\subsection{Detection and Classification}
From the preprocessed video recordings, the 2D positions of the road users in each frame had to be extracted to create trajectories.
As a manual annotation of road users in every frame for the extraction of the trajectories from the images was not feasible, we automated this process.
For this purpose, modern computer vision algorithms based on deep neural networks are well suited. These were also successfully used for the creation of the highD dataset \cite{highD2018}.

Regardless of the chosen architecture of the neural network, it is necessary that an annotated training dataset exists to train the network. Therefore, we annotated road users using polygons in 400~images from all recordings.
Classical augmentation methods such as flipping, rotations and deformations were used to artificially multiply the number of annotated images. Further, Generative Adversarial Networks were used to synthesize additional variations of difficult samples to make the network more robust \cite{vegan}.

Typically, one of two established approaches are used for the detection of objects in images.
Detection networks like SSD \cite{Wei.2016} or Faster-RCNN \cite{Faster} estimate the position and orientation of objects in the form of (rotated) bounding boxes. 
However, this approach is not suitable here because many objects can only imprecisely described from a bird's-eye view by a bounding box. 
For objects such as a moving pedestrian or a bending articulated bus, a more precise description is necessary. 
Thus, we decided to use semantic segmentation, as it is able to assign a class (e.g. car or background) to each pixel. Based on the class information of the individual pixels, a polygon can automatically be derived for each detected object.

Compared to the detection of vehicles on highways \cite{highD2018}, the accurate detection of every kind of road users at intersections proved to be more challenging. 
On the one hand, the size differences between pedestrians and buses are very large.
On the other hand, pedestrians and bicyclists in particular are only described by very few pixels. 
Due to the flight altitude of up to 100~meters and the 4K resolution, each pixel covers a range of about 4x4 centimeters. 
This results in a typical size of a non-moving pedestrian in the image of 10x10~pixels. 
Therefore, we decided to train two separate networks for small and large objects. Both networks are based on the U-Net architecture \cite{unet} and have been adapted to the size of the objects to be detected.

The detections resulted from applying thresholds and morphological operations to the segmented image in order to eliminate outliers and to describe the recorded objects as polygons.
To classify the objects into the classes pedestrian, cyclist, car, truck and bus, it was sufficient to compare the size, speed and position of the respective object with expected values. 
Ambiguous cases, which account for less than 1~\%, were manually annotated.

\begin{table*}[ht!]
\caption{Comparison of existing road user trajectory datasets}
\begin{center}
\begin{tabular}{m{0.14\textwidth} m{0.16\textwidth} m{0.1\textwidth} m{0.08\textwidth} m{0.185\textwidth} m{0.1\textwidth} m{0.07\textwidth} m{0.1\textwidth}}
\cline{1-7} 
\textbf{Title} & \textbf{Location}& \textbf{\# Trajectories}& \textbf{\# Locations}& \textbf{Road User Types} & \textbf{Data Frequency} & \textbf{Method} \\
\hline
\textbf{BIWI Hotel} & sidewalk, hotel entry & 389 & 1 & pedestrian & 2.5 Hz & stat. sensor  \\
\hline
\textbf{BIWI ETH} & university building entry & 360 & 1 & pedestrian & 2.5 Hz & stat. sensor  \\
\hline
\textbf{Crowds UCY/Zara} & campus, urban street & 909 & 3 & pedestrian & 2.5 Hz  & stat. sensor \\
\hline
\textbf{Ko-PER} & urban intersection & 350 & 1 & pedestrian, bicycle, car, truck & 25 Hz & stat. sensor \\
\hline
\textbf{VRU Trajectory} & urban intersection & 3278 & 1 & pedestrian, bicycle & 25 Hz & stat. sensor \\
\hline
\textbf{DUT} & campus & 1862 & 2 & pedestrian, vehicles &  23.98 Hz & drone \\
\hline
\textbf{CITR} & designed experiment & 340 & 1 & pedestrian, golf-cart &  29.97 Hz & drone \\
\hline
\textbf{Stanford Drone} & campus & 10240 & 8 & pedestrian, bicycle, car, skateboard, cart, bus & 25 Hz & drone \\
\hline
\textbf{inD} & \textbf{urban intersection}  & \textbf{11500} & \textbf{4} & \textbf{pedestrian, bicycle, car, truck, bus} & \textbf{25 Hz} & \textbf{drone}  \\
\hline
\end{tabular}
\label{tab1}
\end{center}
\end{table*}

\subsection{Tracking and Post-processing}
As the detection algorithm is run on each frame independently, a tracking algorithm was necessary to connect detections in different frames to tracks.
During this process, detections between two consecutive frames were matched by their distances.
By doing so, false positive detections were completely removed.
If a road user was not detected in a few frames, due to e.g. occlusion by a traffic sign, future positions were predicted until a new detection matched the road user’s track.
After the tracking had been done for the complete recording, additional post-processing was applied to retrieve smooth positions, speeds, and accelerations in both x- and y-direction.
Using Bayesian Smoothing and a constant acceleration model, the trajectories were refined based on information given by all detections belonging to the corresponding track. 
This allowed errors in detection to be replaced by interpolation and improved the positioning error to a pixel size level.
As a final step, all trajectories were converted from image coordinates to metric coordinates and shifted to the same local coordinate system for each recording site.
This was not only necessary, because most applications require the trajectories to be in metric coordinates, but also because the drone did not fly at the exact same position and height at each intersection. 
\subsection{Dataset Format and Tools}
Our goal is to ensure that the dataset is as easy to use as possible. We therefore provide meta data and scripts in addition to the trajectories themselves. 
The dataset is provided in the form of one image and three CSV files per recording. 
While the image shows the intersection from the drone's point of view, the first two CSV files contain meta data about the recording and the tracked road users. 
The meta information includes, for example, the locations, time and duration of the recordings as well as the type, track duration or average speed of each road user. 
The third CSV file contains the trajectories for which the position, orientation, speed and acceleration are given for every frame of the recording.

Matlab and Python source code to import and handle the data, create visualizations and extract maneuvers is provided and constantly updated at https://github.com/ika-rwth-aachen/inD-dataset.

\section{Dataset Statistics and Evaluation}
\subsection{inD at a Glance}
The inD dataset includes trajectories of more than 11~500~road users, which are beside cars, trucks and busses more than 5000~VRUs such as pedestrians and bicyclists.
The inD dataset contains trajectories of vehicles including cars, trucks and busses as well as VRUs such as pedestrians and bicyclists. 
The exact number of trajectories extracted will be given on the website when the dataset is released.

The trajectories are extracted from drone video recordings made at German intersections in Aachen from 2017 to 2019. 
At four different locations, recordings were taken with an typical duration of around 20 minutes covering intersection areas of 80x40~meters to 140x70~meters.
As shown in Fig.~\ref{4fig}, all four intersections are unsignalized, the speed limit is at 50~km/h and walkways exist.
Apart from that, the measurement locations differ in terms of intersection shape, the number and types of lanes, right-of-way rules, traffic composition and kind of interaction.

\begin{figure}[b]
\centerline{\includegraphics[width=\linewidth]{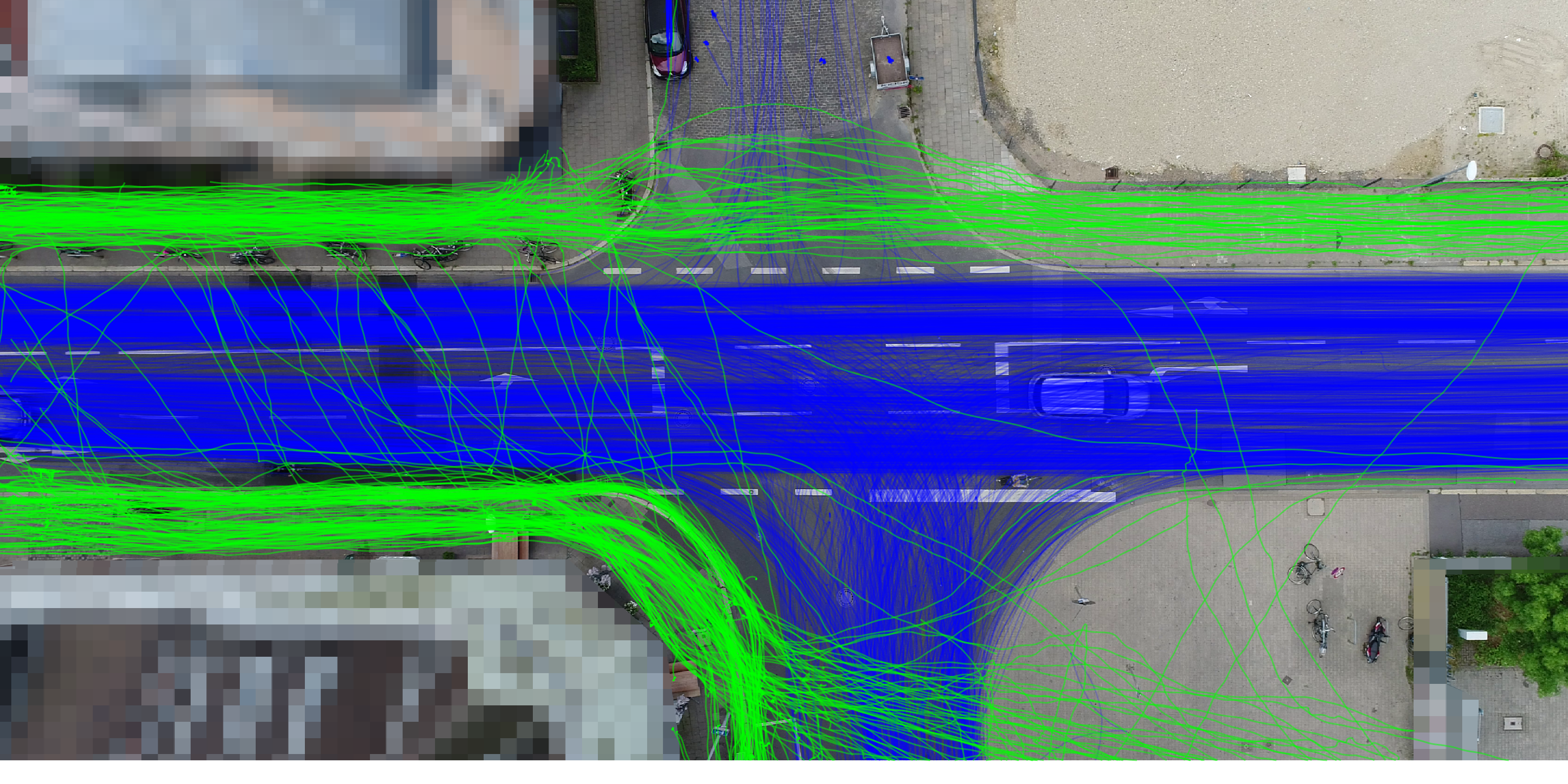}}
\caption{Example trajectories of detected vehicles (blue) and pedestrians (green) at the recording site \textit{Bendplatz}}
\label{figTracks}
\end{figure}

\textit{Bendplatz} is a four-armed intersection with a priority road and is located near a university. There are two left turn lanes, but no regulated pedestrian crossings. 
Due to the proximity to the university, there is an increased frequency of pedestrians, cyclists and buses. 
Between these and the turning vehicles there is the most common interaction at this location. Some of the extracted trajectories of pedestrians and vehicles are visualized in Fig.~\ref{figTracks}.

The intersection at the \textit{Frankenburg} near the city centre has four arms. Directly next to the crossing is a zebra crossing and many parking lots. 
At the intersection the right before left rule applies. 
Due to the location in a residential area and at a park, there is a high number of cyclists and pedestrians. 
While the vehicles and cyclists interact with each other mainly at the intersection, the interaction with pedestrians takes place primarily at the zebra crossing. In addition, traffic is affected by vehicles moving in or out of parking spaces.

The intersection \textit{Heckstrasse} is located in a suburb and is a T-junction. The main road has the right of way and there is a left turn lane into the side road. At the edges of the main road there are cycle paths and in the middle exists a pedestrian crossing. Interactions occur here due to vehicles making turns.

Located in an industrial area next to a bus depot is the T-junction \textit{Neukoellner Strasse}. The priority road is double-laned in both directions and has a left turn area into the side road. The proximity to a motorway access road results in a lot of transit traffic. Because of the bus depots located in the side street, many buses turn off. Further interaction is caused by pedestrians going via a traffic island to one of the two bus stops.

\subsection{Comparison with Existing Datasets}
In Table~\ref{tab1}, we compare the inD dataset with current public datasets, which are the most similar to the inD dataset: Crowds UCY/Zara \cite{Lerner.2007}, Ko-PER \cite{Strigel.2014}, DUT \cite{Yang2019TopviewTA}, CITR \cite{Yang2019TopviewTA}, BIWI Hotel/ETH \cite{SPellegrini.2009}, VRU Trajectory Dataset \cite{goldhammer2018intentions} and Stanford Drone \cite{Robicquet.2016}.
Most of the datasets in this table provide a small amount of trajectories, while the Stanford Drone Dataset and inD contain significantly more road user trajectories.
The Stanford Drone Dataset, however, consists of a large proportion of pedestrians and only a few vehicles, which are mainly parked.
Thus, there are not many interactions between VRUs and vehicles as well as interactions between vehicles. 
Furthermore, inD contains a more representative distribution of road user types on the measured public urban intersections.
By that, inD is much more relevant for applications in the field of automated driving.
Finally, the inD dataset exceeds the Stanford Drone Dataset in accuracy as the inD dataset contains tracks based on pixel-accurate segmentation in images with 4K resolution, while the Stanford Drone Dataset contains tracks based on bounding boxes in images with significantly lower resolution (595 x 326 pixel) \cite{Becker.2018}.

\section{Conclusion}
In this paper we have motivated the need for trajectory data from road users at urban intersections which cannot yet be met by publicly available datasets. 
We have shown that the bird's-eye view of a drone is better suited than other recording methods to record such data. 
After creating a complete processing pipeline, we used that pipeline to create the inD dataset. 
Using deep learning algorithms, we extracted the trajectories of road users such as vehicles and pedestrians with pixel accuracy from 10 hours of video recordings at a total of four recording locations. 
Thus, we surpass any comparable dataset both in size and accuracy, while including a variety of public intersections rather than a university campus.
We will release the dataset upon the conference date.

\bibliographystyle{abbrv}
\bibliography{adiliterature}
\end{document}